\documentclass[lettersize,journal]{IEEEtran}
\usepackage{amsmath,amsfonts}
\usepackage{algorithmic}
\usepackage{algorithm}
\usepackage{array}
\usepackage[caption=false,font=normalsize,labelfont=sf,textfont=sf]{subfig}
\usepackage{textcomp}
\usepackage{stfloats}
\usepackage{url}
\usepackage{verbatim}
\usepackage{graphicx}
\usepackage{cite}
\usepackage{multirow}
\usepackage[colorlinks,linkcolor=blue,anchorcolor=blue,
citecolor=blue]{hyperref}

\usepackage{graphicx}
\usepackage{amsmath,amssymb}
\usepackage{booktabs,tabularx,multirow,threeparttable}
\usepackage{float}
\usepackage{framed}
\usepackage{amssymb}
\usepackage{threeparttable}
\usepackage{multirow}
\usepackage{ragged2e}
\usepackage{enumitem}
\usepackage{makecell}
\usepackage{diagbox}

\usepackage{pifont}
\usepackage[table,xcdraw]{xcolor}
\usepackage{subcaption} 
\usepackage{caption}

\ifCLASSOPTIONcompsoc
  \usepackage[nocompress]{cite}
\else
  \usepackage{cite}
\fi

\usepackage{xcolor}

\begin{document}

\title{T-STAR: A Large-Scale Benchmark for Spatio-Temporal Panoptic Scene Graph Generation in Satellite Video}
\author{
Linlin Wang, Xue Yang,~\IEEEmembership{Member,~IEEE}, Zhihuang Zhou, Zhenyu Zhong, Ruiyuan Zhang, and \\ Yansheng Li,~\IEEEmembership{Senior Member,~IEEE}

\thanks{
\hspace*{\parindent}Linlin Wang, Zhihuang Zhou, Zhenyu Zhong, Ruiyuan Zhang, and Yansheng Li are with School of Remote Sensing and Information Engineering, Wuhan University, Wuhan 430079, China (e-mail: wangll@whu.edu.cn; zhouzhihuang@whu.edu.cn; zhongzy@whu.edu.cn; ruiyuanz@whu.edu.cn; yansheng.li@whu.edu.cn). \\
\hspace*{\parindent}Xue Yang is with School of Automation and Intelligent Sensing, Shanghai Jiao Tong University, Shanghai 200240, China (e-mail: yangxue-2019-sjtu@sjtu.edu.cn). \\
(Corresponding author: Yansheng Li.)}
}




\maketitle

\begin{abstract}
Structured understanding of satellite video is essential for advancing dynamic geospatial scene analysis from low-level perception to high-level cognition. To move beyond object-centric perception, this paper introduces spatio-temporal panoptic scene graph generation (TPSG) in satellite video as a new benchmark task. TPSG aims to generate a structured graph composed of a set of triplets \textit{$<$subject, relationship, object$>$} with explicit temporal spans, thereby describing dynamic geospatial scenes by jointly modeling identity-consistent instance masks and spatio-temporal relationships among panoptic scene elements. However, there is still no dedicated dataset for TPSG in satellite video. Moreover, TPSG in satellite video is intrinsically challenging, as objects are often small and weakly textured, cross-frame association is easily disrupted by occlusion and background clutter, and relationship semantics are highly coupled with spatial structure and temporal evolution. Consequently, TPSG models developed for natural videos are not directly applicable to satellite video. This paper presents T-STAR, a large-scale benchmark dataset for TPSG in satellite video, comprising over 1.1 million instance masks and over 3.8 million spatio-temporal triplets across 39 fine-grained object categories and 70 fine-grained relationship categories. To enable TPSG in satellite video, we propose a unified framework to enhance cross-frame instance consistency and spatio-temporal relationship prediction. Extensive experiments demonstrate the significance of T-STAR and the effectiveness of the proposed framework, establishing a strong benchmark for future research on structured satellite video understanding. The dataset and code are available at \url{https://github.com/linlin-dev/T-STAR}.
\end{abstract}


\section{Introduction}
\IEEEPARstart{S}{atellite} remote sensing has become an important means of Earth observation, benefiting from advances in spaceborne sensing systems that enhance spatial, temporal, and spectral imaging capabilities~\cite{burke2021using,wu2025semantic}. As a dynamic observation modality, satellite video enables persistent monitoring of the Earth's surface, recording not only the spatial distribution of objects but also their state changes, interactions, and scene evolution over time~\cite{zhao2023vehicle,ni2025hazy}. In recent years, substantial progress has been made in satellite video interpretation, including object detection~\cite{xiao2024highly,zhang2021moving}, object segmentation~\cite{zhong2022spatio,li2023multitask}, object tracking~\cite{wang2024mctracker,chen2024satellite}, and scene classification~\cite{gu2020deep,guo2024satellite}. However, these tasks mainly focus on visual perception, such as identifying what is present, where it is located, and how it moves, while falling short of cognitive understanding of complex interactions among objects in satellite video. Spatio-temporal scene graph generation ~\cite{cong2021spatial,ji2020action} provides a structured representation of dynamic geospatial scenes by jointly characterizing object entities, interaction semantics, and their temporal evolution, thereby advancing satellite video interpretation from perception to cognition.

\begin{figure*}[!tb]
	\begin{center}
		\includegraphics[width=1.00\linewidth]{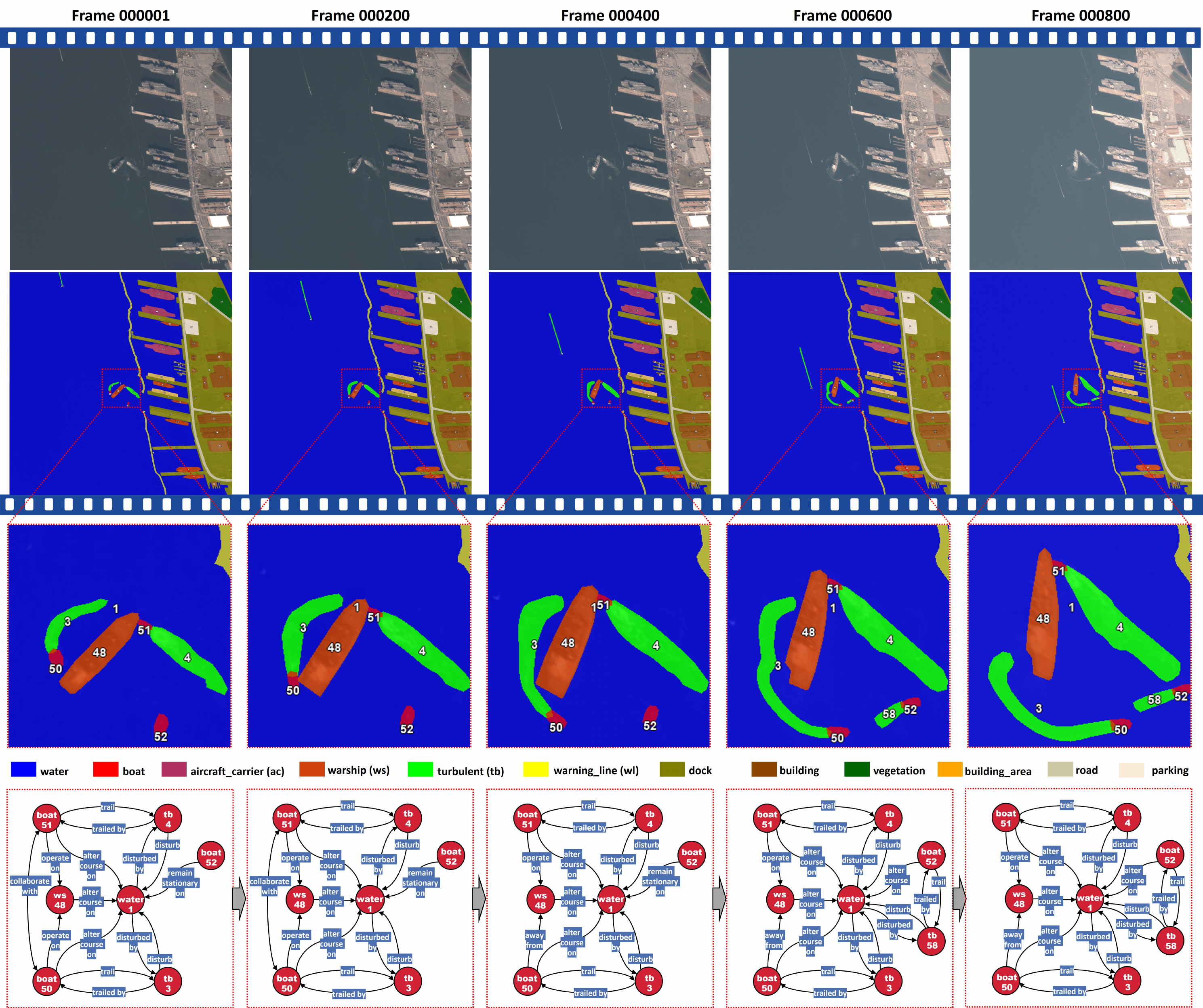}
	\end{center}
	\caption{Illustration of spatio-temporal panoptic scene graph generation (TPSG) in satellite video.}
	\label{fig:TPSG}
\end{figure*}

Recent years have witnessed increasing interest in scene graph generation (SGG) within the computer vision community~\cite{chang2021comprehensive}. The progression from static SGG~\cite{tang2020unbiased,xu2017scene} to box-level spatio-temporal SGG~\cite{nag2023unbiased,wang2024oed}, and further to instance-level spatio-temporal SGG~\cite{yang2023panoptic,nguyen2025motion}, represents an important step toward more comprehensive visual understanding. This trend is particularly relevant to satellite video, where scene semantics are often expressed through dynamic object interactions and long-term relationship evolution. Fig.~\ref{fig:TPSG} illustrates a representative example in a military dock scene. Specifically, \textit{boat50} and \textit{boat51} initially collaborate around \textit{warship48}. As time progresses, \textit{boat50} gradually changes from servicing \textit{warship48} to moving away from it, while the relationship between \textit{boat50} and \textit{boat51} evolves from cooperation to no direct interaction. Meanwhile, the nearby \textit{boat52} changes from being stationary on \textit{water1} to moving on \textit{water1}. 

Although spatio-temporal SGG benchmarks have been developed for natural videos~\cite{ji2020action,yang2023panoptic} and recently extended to aerial videos~\cite{nguyen2024cyclo}, they are not directly applicable to satellite video due to substantial differences in imaging viewpoint, object scale, and scene layout. From the satellite viewpoint, precise mask annotations help reduce background interference and facilitate fine-grained category discrimination, as also highlighted by prior studies~\cite{li2023multitask,chen2025rest,li2026full}. Such representations are particularly important for satellite video, as they support reliable cross-frame association and help preserve identity-consistent instance trajectories. Meanwhile, rich spatio-temporal relationships arise among both dynamic and static scene elements, including natural geographic elements, man-made facilities, transportation objects, and accompanying wake-like regions. Motivated by these characteristics, this paper introduces spatio-temporal panoptic scene graph generation (TPSG) in satellite video as a new benchmark task for structured satellite video understanding. Nevertheless, there remains a significant lack of dedicated datasets and approaches for TPSG in satellite video.

As summarized in Table~\ref{tab:dataset_comparison}, existing satellite video datasets are mainly designed for object detection, segmentation, or tracking. Although their annotation granularity and annotation volume have been continuously improved, they still lack relationship annotations and usually provide bounding-box annotations that inevitably include background regions, while mainly focusing on coarse-grained transportation objects. Given these limitations, it is impractical to directly build a TPSG benchmark on top of these datasets. Specifically, some characteristics should be considered when annotating triplets for TPSG in satellite video: (i) \textbf{Fine-grained objects with temporally consistent instance-level annotations}. Fine-grained objects in satellite video often exhibit only subtle inter-category differences in contour, and thus should be annotated with temporally consistent instance-level masks to accurately delineate object boundaries and maintain cross-frame identity consistency; (ii) \textbf{Panoptic scene elements spanning diverse dynamic geospatial scenarios}. All relevant scene elements in satellite video should be covered, including natural geographic elements, such as water and vegetation, man-made facilities, such as aprons and terminals, transportation objects, such as airplanes and ships, and accompanying wake-like regions, such as Kelvin wakes and turbulent wakes, so that they can be jointly modeled in a panoptic manner; and (iii) \textbf{High-value relationships with explicit spatio-temporal characteristics}. Object pairs and their relationships should be annotated in a one-to-many way by jointly considering spatial and temporal evidence, so that both spatial relationships and spatio-temporally evolving interactions can be explicitly represented.

\begin{table*}[t!]
\small
\centering
\caption{Comparison between the T-STAR dataset and representative satellite video datasets. HBB denotes horizontal bounding box, OBB denotes oriented bounding box, Mask denotes mask annotations for transportation objects only, and Panoptic Mask denotes panoptic annotations over all relevant scene elements, including natural geographic elements, man-made facilities, transportation objects, and accompanying wake-like regions.}
\renewcommand{\arraystretch}{1.8}
\resizebox{1.0\textwidth}{!}{
\begin{tabular}{cccccccc}
\hline
\multirow{2}{*}{\rule{0pt}{3.6ex}\textbf{Datasets}} &
\multirow{2}{*}{\rule{0pt}{3.6ex}\textbf{Annotation}} &
\multirow{2}{*}{\rule{0pt}{3.6ex}\textbf{Image size}} &
\multicolumn{2}{c}{\textbf{Object}} &
\multicolumn{2}{c}{\textbf{Relationship}} &
\multirow{2}{*}{\rule{0pt}{3.6ex}\textbf{Venue}} \\
\cline{4-7}
 &  & 
 & \begin{tabular}[c]{@{}c@{}}Instances\end{tabular}
 & \begin{tabular}[c]{@{}c@{}}Categories\end{tabular}
 & \begin{tabular}[c]{@{}c@{}}Triplets\end{tabular}
 & \begin{tabular}[c]{@{}c@{}}Categories\end{tabular}
 &  \\
\hline
VISO~\cite{yin2021detecting} & HBB & 220$\times$223$\sim$1,348$\times$1,348 & 1,646,038 & 4 & - & - & TGRS'21 \\
SatSOT~\cite{zhao2022satsot} & HBB & 499$\times$335$\sim$1,298$\times$1,376 & 27,664 & 4 & - & - & TGRS'22 \\
SV248S~\cite{li2022deep} & Mask & 213$\times$312$\sim$827$\times$971 & 163,234 & 4 & - & - & GRSM'22 \\
AIR-MOT~\cite{he2022multi} & HBB & 1,920$\times$1,080 & 5,736 & 2 & - & - & TGRS'22 \\
SAT-MTB~\cite{li2023multitask} & HBB/OBB/Mask & 502$\times$512$\sim$3,000$\times$1,500 & 1,033,511 & 14 & - & - & TGRS'23 \\
TMS~\cite{zhao2023vehicle} & HBB & 512$\times$512$\sim$540$\times$480 & 128,801 & 1 & - & - & TPAMI'24 \\
OOTB~\cite{chen2024satellite} & OBB & 142$\times$136$\sim$3,000$\times$1,500 & 29,890 & 4 & - & - & ISPRS'24 \\
LMOD~\cite{zhang2025structural} & HBB & 1,500$\times$1,160$\sim$4,000$\times$2,000 & 480,332 & 4 & - & - & J-STARS'25 \\
\hline
T-STAR (Ours) & Panoptic Mask & 400$\times$400$\sim$1,500$\times$1,500 & 1,188,193 & 39 & 3,832,449 & 70 & - \\
\hline
\end{tabular}}
\label{tab:dataset_comparison}
\end{table*}

To address the scarcity of dedicated datasets, this paper presents
\textbf{T-STAR}, a large-scale benchmark dataset for TPSG in satellite
video. In this dataset, (i) satellite videos with a spatial resolution
of approximately 1~m are collected, covering representative
activity-intensive geospatial scenarios, including maritime, bridge,
port, airport, and railway scenes; (ii) under the guidance of domain experts, panoptic scene elements are categorized into 39 fine-grained classes and panoptically annotated with temporally consistent instance-level masks, while all relationships are annotated with 70 fine-grained categories; and (iii) valid relationships between object pairs are annotated in a one-to-many
manner, where each pair may be associated with multiple relationship
categories over time, and all relationship categories are defined in a
rotation-invariant manner. In summary, T-STAR provides significant advantages over existing satellite video datasets mainly designed for detection or tracking, and establishes a richer benchmark for structured understanding of dynamic geospatial scenes.

Considering the unique characteristics of satellite video, effective TPSG models should address two fundamental challenges. First, objects in satellite video are often small and weakly textured, while occlusion, background clutter, and appearance ambiguity can easily disrupt cross-frame association, making it difficult to construct reliable identity-consistent instance trajectories. Second, spatio-temporal relationships in satellite video are not isolated static labels but dynamic semantics jointly shaped by spatial configurations and temporal evolution. TPSG therefore requires models to capture structured dependencies among instance pairs within each frame and relationship evolution across frames, especially for high-value relationships involving persistent interactions, relative motion patterns, and long-term state transitions. To address these challenges, we propose a unified framework for the TPSG task in satellite video. First, a video panoptic parsing model is employed to extract initial instance features. Then, a spatio-temporal cooperative learning (STCL) model is designed for spatio-temporal relationship prediction, where memory-guided matching (MGM) enhances cross-frame instance consistency, spatial context enhancement (SCE) propagates contextual information among instance pairs, and multi-scale temporal learning (MTL) captures temporal dependencies.

In summary, the main contributions of this paper are as follows:
\begin{itemize}
    \item We present T-STAR, to the best of our knowledge, the first large-scale benchmark dataset for TPSG in satellite video. T-STAR contains over 1.1 million instance masks and over 3.8 million spatio-temporal triplets across 39 fine-grained object categories and 70 fine-grained relationship categories, providing a challenging and practical testbed for structured satellite video understanding.

    \item To enhance TPSG in satellite video, we propose a spatio-temporal cooperative learning (STCL) model that coordinates memory-guided matching (MGM), spatial context enhancement (SCE), and multi-scale temporal learning (MTL), thereby improving cross-frame instance consistency and spatio-temporal relationship prediction.

    \item Extensive experiments on T-STAR demonstrate the effectiveness of the proposed framework under both PredCls and SGDet settings, establishing a solid experimental foundation for future research on TPSG in satellite video.
\end{itemize}

\section{Related Work}

\subsection{SGG in Natural Images and Videos}

Scene graph generation (SGG) in images~\cite{lu2016visual,deng2022hierarchical} aims to identify entities and predict their relationships, forming a structured graph composed of triplets \textit{$<$subject, relationship, object$>$}. Owing to its strong capability in semantic abstraction and structured reasoning, SGG has been widely applied to scene understanding and reasoning tasks~\cite{johnson2015image,yang2019auto,Li_2019_ICCV}. Early studies mainly focused on static natural images, where large-scale annotations have promoted rapid progress in image-level SGG~\cite{sadeghi2011recognition,krishna2017visual,hudson2019gqa,kuznetsova2020open}. Most existing SGG methods follow a two-stage pipeline~\cite{tang2020unbiased,yang2022panoptic}, where visual entities are first identified and pair-level relationships are then predicted. This paradigm has been further advanced through various strategies, including multimodal features~\cite{liang2018visual,dong2022stacked}, prior information~\cite{dai2017detecting,zellers2018neural,hwang2018tensorize}, commonsense knowledge~\cite{yu2017visual,zareian2020bridging}, message passing~\cite{xu2017scene,li2018factorizable,li2021bipartite}, and debiasing~\cite{Sun2023Unbiased, li2024fine, liu2025causal}.

Compared with static images, videos provide continuous observations over time and therefore enable the modeling of object motion, interaction processes, and relationship evolution. This has motivated spatio-temporal SGG~\cite{chen2025diffvsgg,zhang2023end,xu2022meta}, where object trajectories are typically constructed by associating frame-level detections, and relationships are then predicted over object tracklets or trajectories. Traditional studies mainly focus on box-level spatio-temporal SGG~\cite{pu2023spatial,lin2024td2}. Considering the coarse localization of bounding boxes and their limited ability to represent irregular object shapes, recent studies~\cite{yang2023panoptic,nguyen2025motion} have explored mask-level spatio-temporal SGG to enable finer-grained modeling of both countable objects and background regions. Despite these advances, existing spatio-temporal SGG methods are mainly developed for natural videos captured from ground-level or egocentric perspectives. In such scenarios, objects usually occupy relatively large image regions, contain rich appearance cues, and exhibit relationship patterns dominated by human-object or object-object interactions. However, these assumptions do not directly hold for satellite video. Satellite videos are captured from top-down viewpoints, where objects are often extremely small, weakly textured, densely distributed, and easily confused with background clutter. Therefore, it is insufficient to directly transfer spatio-temporal SGG methods from natural videos to satellite video.

\subsection{SGG in Remote Sensing Images and Videos}

With the increasing demand for structured geospatial understanding, SGG has also attracted growing attention in remote sensing~\cite{li2021semantic}. Representative datasets, including GRTRD~\cite{chen2021message}, RSSGD~\cite{li2021semantic}, S2SG~\cite{lin2022srsg}, STAR~\cite{li2025star}, and ReCon1M~\cite{yan2025recon1m}, together with recent methods based on prototype learning~\cite{hou2025scene,wang2025hierarchical} and bias-aware learning~\cite{wang2026bias}, have advanced remote sensing interpretation from object-level perception to relationship-level scene cognition. These efforts demonstrate the potential of scene graphs for structured interpretation of complex remote sensing scenes. However, existing remote sensing SGG studies mainly focus on static imagery, where relationships are modeled within a single image. As a result, they cannot capture cross-frame instance consistency or relationship evolution over time. This limitation is particularly critical for satellite video, where high-value relationships often depend on persistent interactions and long-term state transitions.

In contrast to static satellite imagery, satellite video provides continuous observations of surface object movements and state changes over time, and has therefore become increasingly important for dynamic geospatial interpretation~\cite{gu2020detection,li2023recent,wang2026dual}. Existing studies on satellite video interpretation mainly focus on object-centric perceptual tasks~\cite{lai2023target,chen2024satellite}, such as object detection, segmentation, and tracking. Although these efforts provide essential foundations for satellite video interpretation, most existing datasets remain centered on transportation objects, lacking comprehensive coverage of panoptic scene elements, explicit relationship annotations, and unified modeling of instance trajectories, relationships, and temporal spans. Recently, the UAV-based AeroEye dataset has been constructed for box-level spatio-temporal SGG in aerial videos, and CYCLO has been proposed to model multi-object relationships from aerial-oblique views~\cite{nguyen2024cyclo}. Nevertheless, aerial videos still differ substantially from satellite video in imaging viewpoint, object scale, scene organization, and relationship patterns. Satellite videos usually cover broader geographic areas, contain smaller objects with weaker visual textures, and exhibit stronger dependence on geospatial structures and long-term temporal evolution. Therefore, it is necessary to investigate TPSG specifically for satellite videos, with explicit consideration of their unique imaging characteristics, object properties, and relationship patterns. 


\begin{figure*}[!tb]
	\begin{center}
		\includegraphics[width=1.00\linewidth]{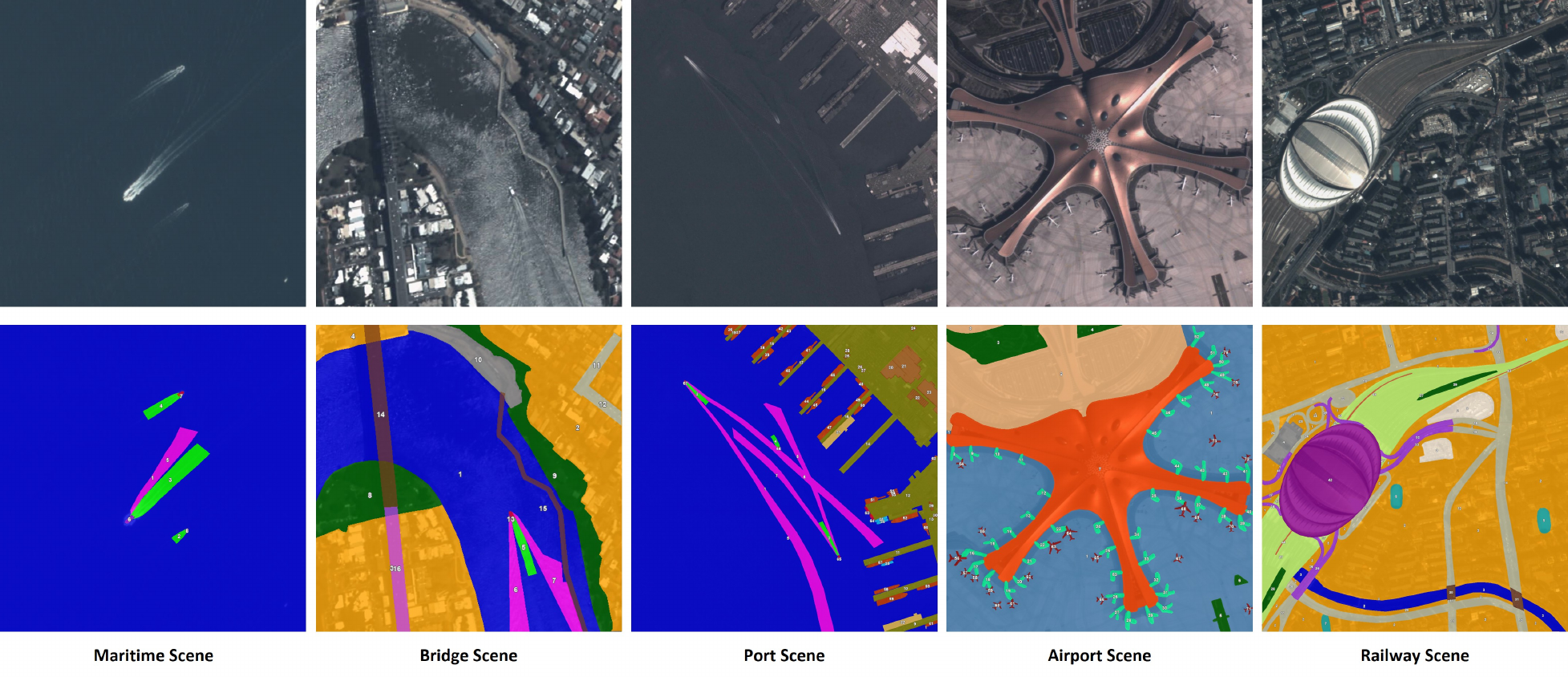}
	\end{center}
	\caption{Representative annotated scenes from the T-STAR dataset.}
	\label{fig:tstar_scenes}
\end{figure*}

\begin{figure*}[!tb]
	\begin{center}
		\includegraphics[width=1.00\linewidth]{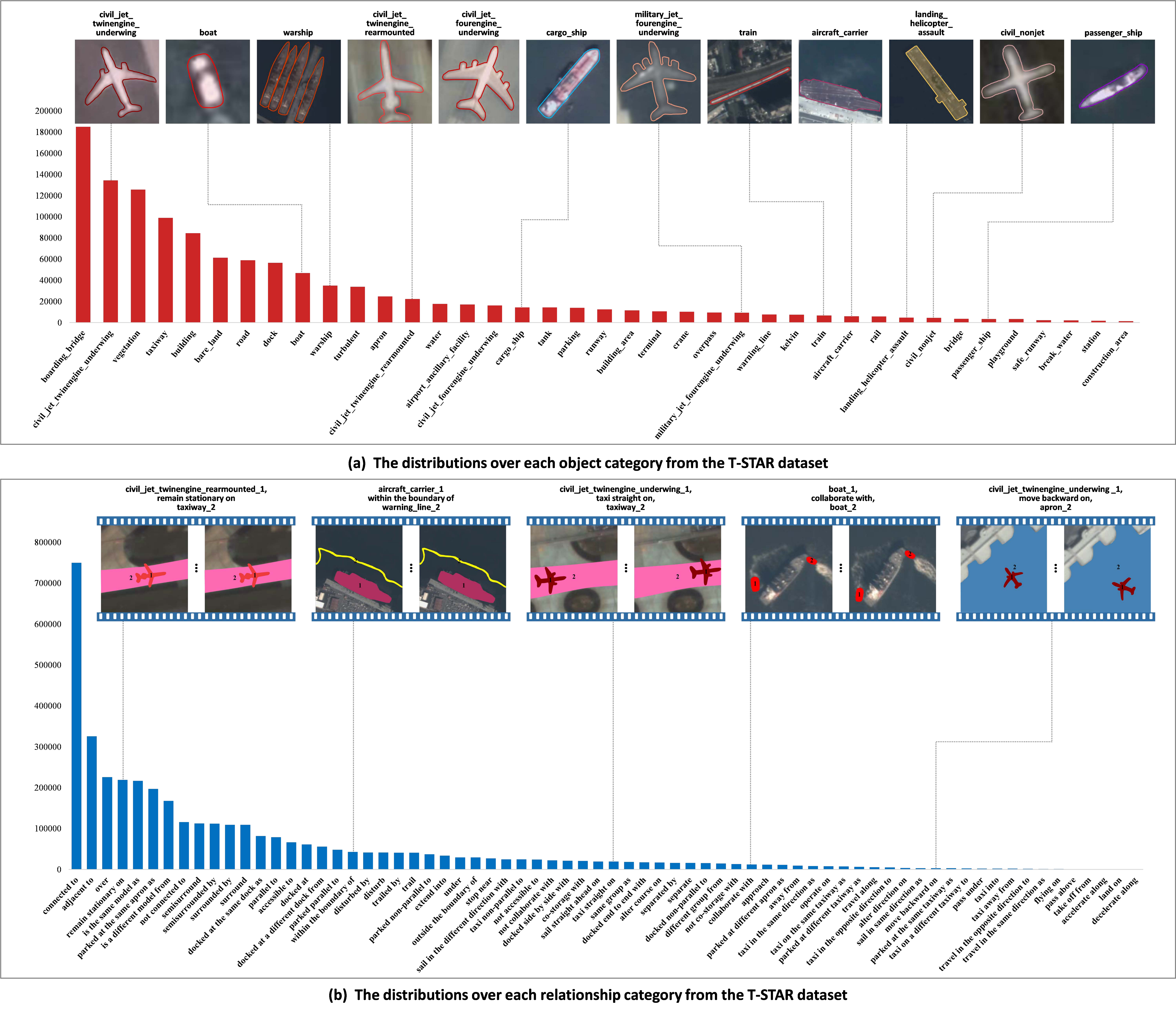}
	\end{center}
	\caption{Statistics and visualization of objects (a) and relationships (b) from the T-STAR dataset.}
	\label{fig:object+relationships}
 \vspace{-8pt}
\end{figure*}

\begin{table}[t]
\centering
\small
\caption{Overview of the T-STAR dataset.}
\label{tab:dataset_overview}
\renewcommand{\arraystretch}{1.3}
\setlength{\tabcolsep}{6pt}
\begin{tabular}{l|l}
\hline
\textbf{Item} & \textbf{Statistics} \\
\hline
Data source & Jilin-1 satellite video \\
Spatial resolution & Approximately 1 m \\
Image size & 400$\times$400$\sim$1,500$\times$1,500 pixels \\
Number of videos & 150 \\
Number of frames & 34,333 \\
Frames per video & 49$\sim$848 (avg. 228.9) \\
Object categories & 39 \\
Relationship categories & 70 \\
Instance masks & 1,188,193 \\
Spatio-temporal triplets & 3,832,449 \\
Train/val/test split & 6:2:2 at the video level \\
\hline
\end{tabular}
\end{table}

\section{Details of the T-STAR Dataset}
\subsection{Data Sources and Scenario Selection}

T-STAR is mainly constructed from Jilin-1 satellite video, including both 
publicly accessible and commercially acquired videos. These data provide meter-level spatial resolution and continuous frame observations, making them suitable for capturing fine-grained objects, maintaining cross-frame instance consistency, and modeling spatio-temporal relationships in satellite video. Since many geospatial scenes exhibit limited temporal variation within short video durations, we deliberately select activity-intensive dynamic geospatial scenarios with rich and observable interaction processes. Specifically, T-STAR covers representative scene types, including maritime, bridge, port, airport, and railway scenes. To provide a clearer overview of the dataset scale and video-level properties, Table~\ref{tab:dataset_overview} summarizes the main metadata of T-STAR, including the number of videos, frame statistics, image sizes, annotation categories, and dataset splits. These statistics indicate that T-STAR provides large-scale frame-level annotations and diverse video-level temporal observations, making it suitable for evaluating TPSG models under realistic satellite video conditions.

Fig.~\ref{fig:tstar_scenes} shows typical annotated examples of these scenes. 
Airport scenes contain functionally organized regions such as runways, taxiways, aprons, terminals, and boarding bridges, where airplane operations naturally involve temporally evolving interaction semantics. Port and maritime scenes contain ships, docks, breakwaters, cranes, and surrounding buildings, leading to flexible motion patterns and diverse spatio-temporal relationships. Bridge and railway scenes further enrich the dataset with bridges, trains, stations, rails, water areas, and surrounding infrastructure, where relationships are jointly shaped by spatial connectivity, traffic movement, and temporal operation processes. Overall, these scenes exhibit clear spatial organization patterns, diverse object compositions, and abundant observable dynamic interactions, making them particularly suitable for TPSG research.

\subsection{Annotation Taxonomy Design}

\subsubsection{Object Taxonomy}

T-STAR defines a fine-grained object taxonomy consisting of 39 categories, including natural geographic elements, man-made facilities, transportation 
objects, and accompanying wake-like regions. This taxonomy is designed based on key components in typical satellite video scenes, with explicit consideration of cross-frame traceability and category discriminability. For airport scenes, it includes fine-grained airplane categories, \textit{runways}, \textit{taxiways}, \textit{aprons}, \textit{terminals}, \textit{boarding\_bridges}, and surrounding geographic elements and facilities; for port and maritime scenes, it covers fine-grained ship categories, \textit{docks}, \textit{cranes}, \textit{breakwaters}, and surrounding scene elements. Accompanying wake-like regions, represented by the categories \textit{kelvin} and \textit{turbulent} in our taxonomy, are also annotated because they provide meaningful semantic cues for understanding dynamic maritime activities. Fig.~\ref{fig:object+relationships}(a) shows representative object categories and their statistics in T-STAR.

\begin{figure*}[!tb]
	\begin{center}
            \includegraphics[width=0.98\linewidth]{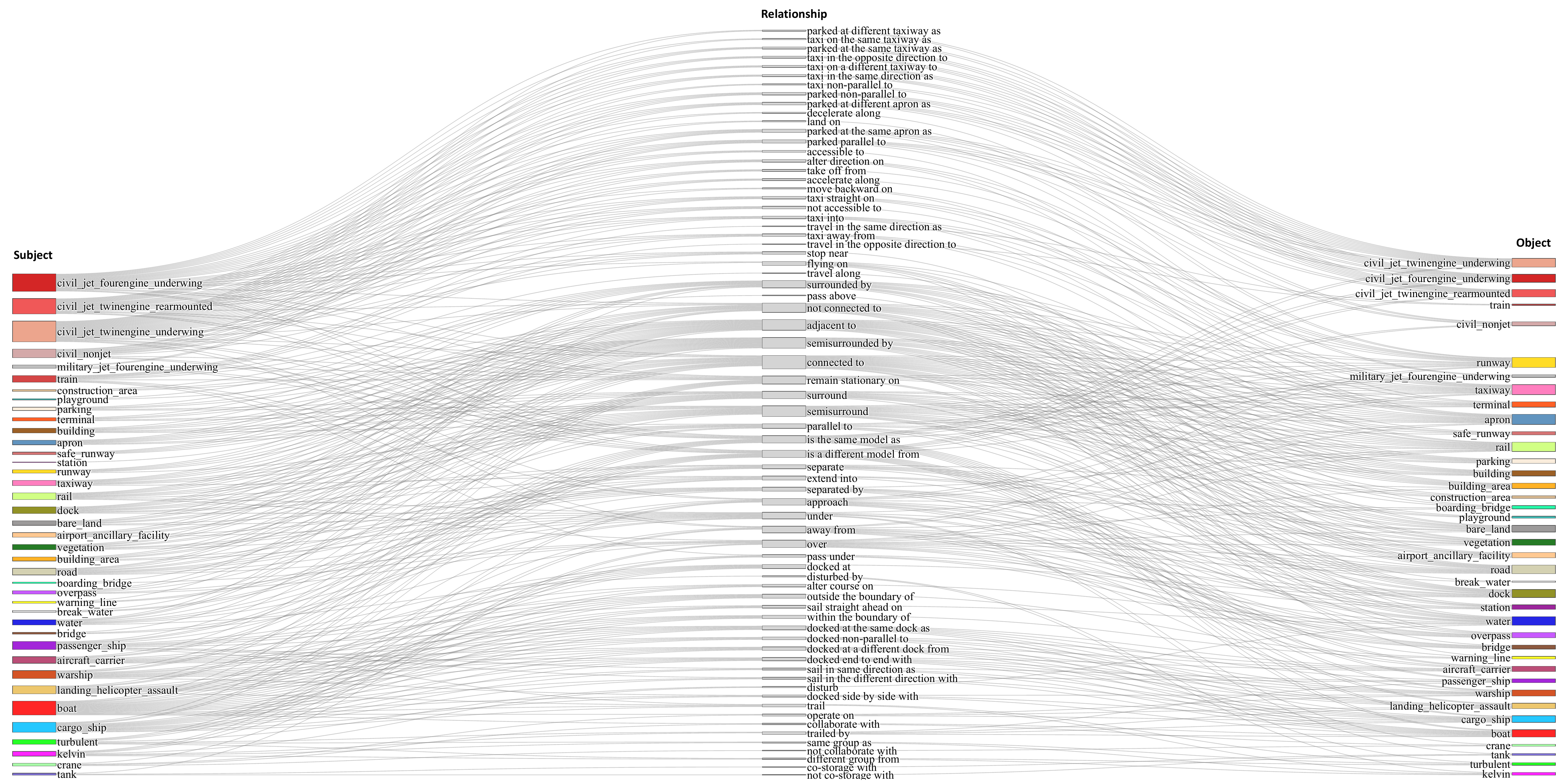}
	\end{center}
	\caption{Interaction mapping among subject categories, relationship categories, and object categories.}
	\label{fig:interaction}
\end{figure*}

A key design principle of T-STAR is that it does not restrict annotations to transportation objects. Instead, all relevant scene elements in each frame are panoptically annotated. This design is essential for TPSG, since high-value relationships in satellite video arise not only among dynamic scene elements, but also between dynamic and static scene elements, as well as among static scene elements. Moreover, fine-grained categorization is particularly important for transportation objects. As shown in Fig.~\ref{fig:object+relationships}(a), airplanes and ships in satellite video often exhibit subtle structural differences while providing weak appearance cues, making cross-frame identity maintenance challenging. By introducing semantically meaningful fine-grained categories, T-STAR reduces category ambiguity and provides stronger semantic constraints for downstream relationship reasoning.

\subsubsection{Relationship Taxonomy}

T-STAR establishes a fine-grained spatio-temporal relationship taxonomy with 70 relationship categories. This taxonomy is designed to jointly characterize stable spatial structures and temporally evolving interaction semantics in satellite video. The relationships are organized into seven major groups: spatial topology, connectivity interaction, carrier interaction, motion interaction, static interaction, identity discrimination, and functional interaction. Specifically, spatial topology relationships, such as \textit{adjacent to}, \textit{within the boundary of}, and \textit{surround}, describe geometric and topological configurations among scene elements. Connectivity interaction relationships, such as \textit{connected to}, \textit{accessible to}, and \textit{not accessible to}, capture structural accessibility and connection states in geospatial scenes. Carrier interaction relationships, such as \textit{taxi on}, \textit{sail on}, and \textit{docked at}, describe carrier-dependent interactions between transportation objects and their underlying scene elements. Motion interaction relationships, such as \textit{approach}, \textit{away from}, and \textit{trail}, characterize dynamic relative motion patterns between instances over time. Static interaction relationships, such as \textit{parked parallel to}, \textit{docked side by side with}, and \textit{stop near}, describe temporally stable interaction states among objects. Identity discrimination relationships, such as \textit{is the same model as}, \textit{same group as}, and \textit{different group from}, distinguish semantic or group-level identity associations. Functional interaction relationships, such as \textit{operate on}, \textit{collaborate with}, and \textit{co-storage with}, further extend relationship modeling from geometric and motion-level descriptions to higher-level scene understanding. Fig.~\ref{fig:object+relationships}(b) presents representative relationship categories and their statistics in T-STAR.

A key characteristic of the relationship taxonomy is its temporal grounding and visual observability. Each annotated relationship must be supported by visual evidence from continuous frame observations, rather than inferred solely from prior knowledge or single-frame spatial layouts. Therefore, relationship modeling in T-STAR is built upon temporally consistent instance trajectories and their motion evolution, enabling both static spatial relationship understanding and process-level spatio-temporal reasoning. In addition, T-STAR allows one-to-many relationship annotations for each trajectory pair, which enriches the structural complexity of the generated scene graphs and enables stable spatial relationships and temporally evolving interactions to be represented simultaneously.

\subsection{Annotation Protocol and Quality Control}

T-STAR adopts a unified annotation protocol to ensure accurate panoptic masks, stable cross-frame identities, and reliable spatio-temporal relationship annotations. The protocol focuses on three aspects: pixel-level annotation quality, cross-frame identity continuity, and the semantic validity of spatio-temporal relationships. Specifically, T-STAR employs frame-by-frame panoptic annotation, where every pixel in each frame is assigned to a predefined dynamic or static scene element category, rather than annotating only transportation objects. Each visible instance is assigned a mask, and the same instance is maintained with a unique identity across frames. For continuously observable instances, mask contours are required to evolve smoothly over time; for partially occluded or weakly visible instances, temporal cues from neighboring frames are used to preserve identity continuity. This protocol naturally forms mask trajectories for all relevant instances, providing a reliable basis for both video panoptic parsing and TPSG.

For relationship annotation, T-STAR annotates spatio-temporal triplets on top of identity-consistent trajectory pairs and explicitly records the temporal span during which each relationship holds. Each triplet is required to be visually verifiable within its corresponding temporal interval, ensuring that relationship labels are grounded in observable spatio-temporal evidence. Relationship annotation follows a one-to-many principle, meaning that a single instance pair may correspond to multiple valid relationships over time. Structural and topological relationships are determined according to relative position, mask overlap, and spatial inclusion. Motion-process relationships are judged based on trajectory trends, relative movement, and speed variation, while static and functional relationships are determined jointly by spatial structure and temporal persistence. This annotation strategy ensures that the same relationship category is interpreted consistently across different scenes and video sequences.

To ensure annotation quality, T-STAR adopts a closed-loop workflow of 
initial annotation--review--revision, assisted by expert guidance 
and systematic consistency checking. Specifically, four trained annotators 
first performed the annotations according to the unified protocol, while two 
domain experts cross-checked the annotation results and reviewed ambiguous 
cases. Potential errors and inconsistent annotations were returned for revision. The quality-control procedure includes object category validity checking, mask boundary accuracy checking, cross-frame identity continuity checking, relationship category validity checking, and temporal-span rationality checking. Through this multi-stage process, T-STAR maintains annotation stability at scale and provides reliable support for benchmark evaluation.

\subsection{Dataset Characteristics and Challenges}

\subsubsection{Long-Tailed Object and Relationship Distributions}

As shown in Fig.~\ref{fig:object+relationships}, both the object and relationship annotations in T-STAR follow pronounced long-tailed distributions across 39 object categories and 70 relationship categories. Frequent categories occupy a large proportion of the annotations, whereas many fine-grained transportation objects and semantically important relationships are sparsely represented. This phenomenon reflects the natural semantic imbalance of real satellite video scenes, but also poses a major challenge for model learning, since models tend to overfit dominant categories and struggle with rare yet critical categories. This challenge is further amplified by the coexistence of transportation objects and diverse scene elements, as well as the one-to-many relationship annotations over trajectory pairs, which together make the semantic composition of T-STAR considerably more complex than that of conventional object-centric satellite video datasets.

\subsubsection{Fine-Grained Semantic Discrimination}

T-STAR contains many fine-grained object categories whose visual differences are subtle under satellite video observations. This is especially evident for airplanes and ships. Different airplane categories may differ mainly in wing layout, engine number, fuselage size, or tail structure, while different ship categories are often distinguished by hull shape, deck layout, functional equipment, or relative size. Such fine-grained distinctions are difficult to preserve under meter-level satellite video observations, thereby increasing both category ambiguity and cross-frame identity drift. In addition, weak texture, low spatial resolution, and background clutter further reduce the discriminability of instance appearances, making it challenging for models to maintain stable identities and infer accurate relationships over time.

\begin{figure*}[!tb]
	\begin{center}
		\includegraphics[width=0.98\linewidth]{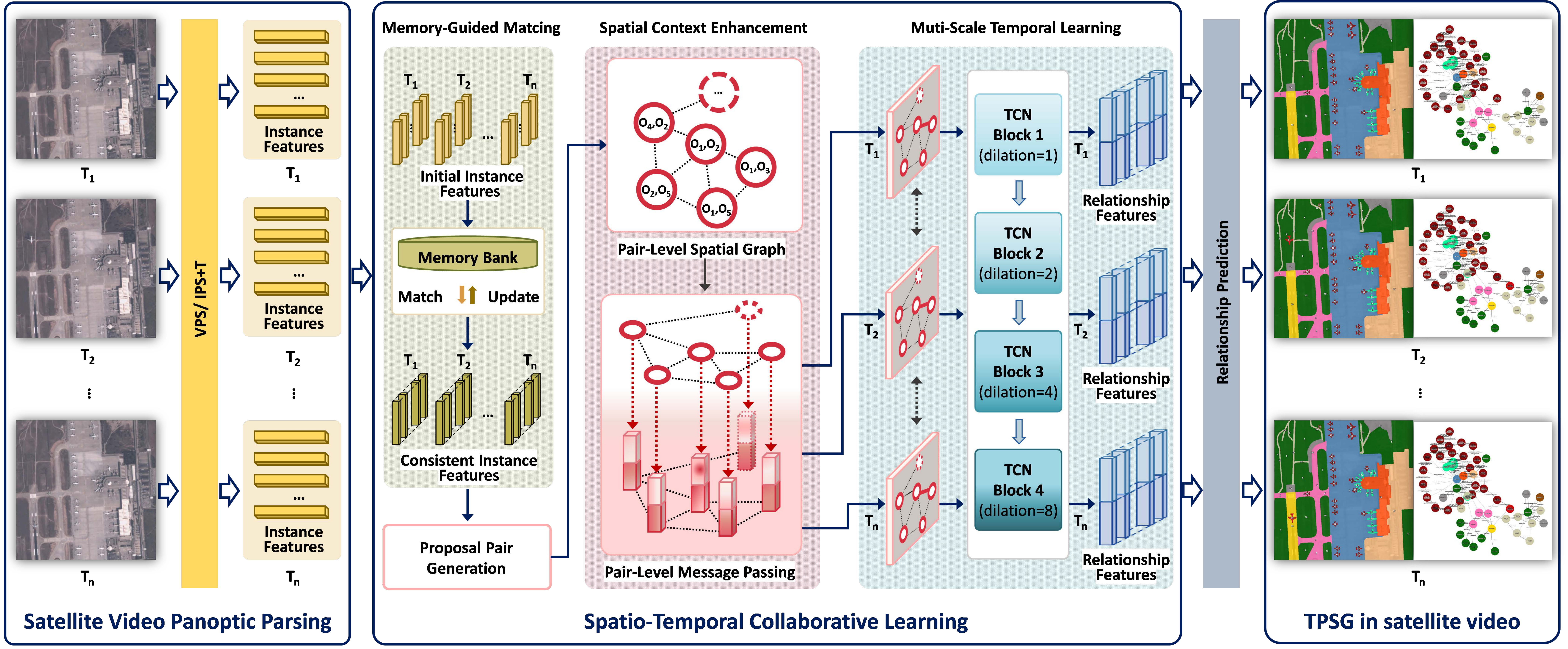}
	\end{center}
    \caption{Overall framework for the TPSG task. Given a satellite video sequence, instance features are obtained via video panoptic parsing. Then, spatio-temporal relationships are predicted through spatio-temporal cooperative learning. Finally, the spatio-temporal panoptic  scene graph is generated.}
	\label{fig:pipeline}
\end{figure*}

\subsubsection{Scale Variation and  Motion Heterogeneity}

Objects in T-STAR show substantial differences in both spatial scale and temporal dynamics. Natural geographic elements and man-made facilities often occupy large image regions and remain structurally stable over time, with only slight apparent motion caused by platform jitter or imaging variations. In contrast, transportation objects and accompanying wake-like regions are usually much smaller, but exhibit clearer displacement, pose variation, and interaction changes across frames. This coexistence of large static scene elements and small transportation objects leads to pronounced scale imbalance and motion heterogeneity. These characteristics increase the difficulty of instance parsing, feature learning, and cross-frame association. Models must remain robust to objects with different spatial extents, visual details, motion patterns, and temporal persistence, while also avoiding confusion between small transportation objects and complex background structures.

\subsubsection{Strong Spatio-Temporal Coupling of Relationship Semantics}

A large portion of the relationships in T-STAR exhibit strong spatio-temporal coupling. Unlike static image scene graphs, many relationship categories in T-STAR cannot be determined from a single frame alone. Their validity depends jointly on spatial structure, trajectory continuity, relative motion trends, and temporal persistence. For example, relationships such as \textit{taxi on the same taxiway as}, \textit{taxi on a different taxiway to}, \textit{docked at the same dock as}, and \textit{docked at a different dock from} require simultaneous verification of spatial configuration and temporal evolution. More broadly, many high-value relationships in T-STAR are not merely frame-wise geometric descriptions, but are instead process-level semantics that emerge, persist, and evolve over time. This makes relationship reasoning in T-STAR inherently more complex than frame-level spatial relationship classification. Fig.~\ref{fig:interaction} further illustrates the structural complexity of object--relationship interactions in T-STAR.

\section{The Proposed Method}
\subsection{Task Formulation and Framework}
Given a satellite video sequence $\mathcal{V}=\{I_1,I_2,\ldots,I_T\}$ 
with $T$ frames, the goal of TPSG is to generate a structured 
spatio-temporal panoptic scene graph $\mathcal{G}_{\mathcal{V}}$. 
The graph consists of a set of identity-consistent triplets $\mathcal{Y}$, 
where each triplet is represented as  $\langle M_S, r_{\Delta t}, M_O \rangle$. Here, $M_S$ and $M_O$ denote the subject and object mask trajectories,
respectively, and $r_{\Delta t}$ denotes a temporally grounded relationship predicate, where $r$ is the relationship category and
$\Delta t$ denotes the temporal span over which the relationship holds. The temporal span may consist of a single interval or multiple disjoint intervals. Formally, the TPSG task in satellite video can be factorized as:

\begin{equation}
P(\mathcal{G}_V \mid \mathcal{V}) =
P(\mathcal{M} \mid \mathcal{V}) \,
P(\mathcal{Y} \mid \mathcal{M}, \mathcal{V}),
\label{eq:TPSG}
\end{equation}
where $\mathcal{M}$ denotes the set of identity-consistent instances across 
all frames, and $\mathcal{Y}$ denotes the set of predicted spatio-temporal triplets constructed on top of $\mathcal{M}$. The first term corresponds to 
video panoptic parsing for extracting temporally consistent instance masks, 
while the second term corresponds to spatio-temporal relationship prediction 
conditioned on the parsed instances and the input video.

As illustrated in Fig.~\ref{fig:pipeline}, the overall framework for the TPSG task in satellite video adopts a two-stage pipeline. In Stage I, instance features are extracted via video panoptic parsing, which can be performed using two baselines~\cite{yang2023panoptic}: image panoptic segmentation with tracker (IPS+T) and video panoptic segmentation (VPS). In Stage II, spatio-temporal relationships are predicted through spatio-temporal cooperative learning (STCL), which comprises three key modules: a memory-guided matching (MGM) module, a spatial context enhancement (SCE) module, and a multi-scale temporal learning (MTL) module.

\begin{figure}[!tb]
	\begin{center}
		\includegraphics[width=1.0\linewidth]{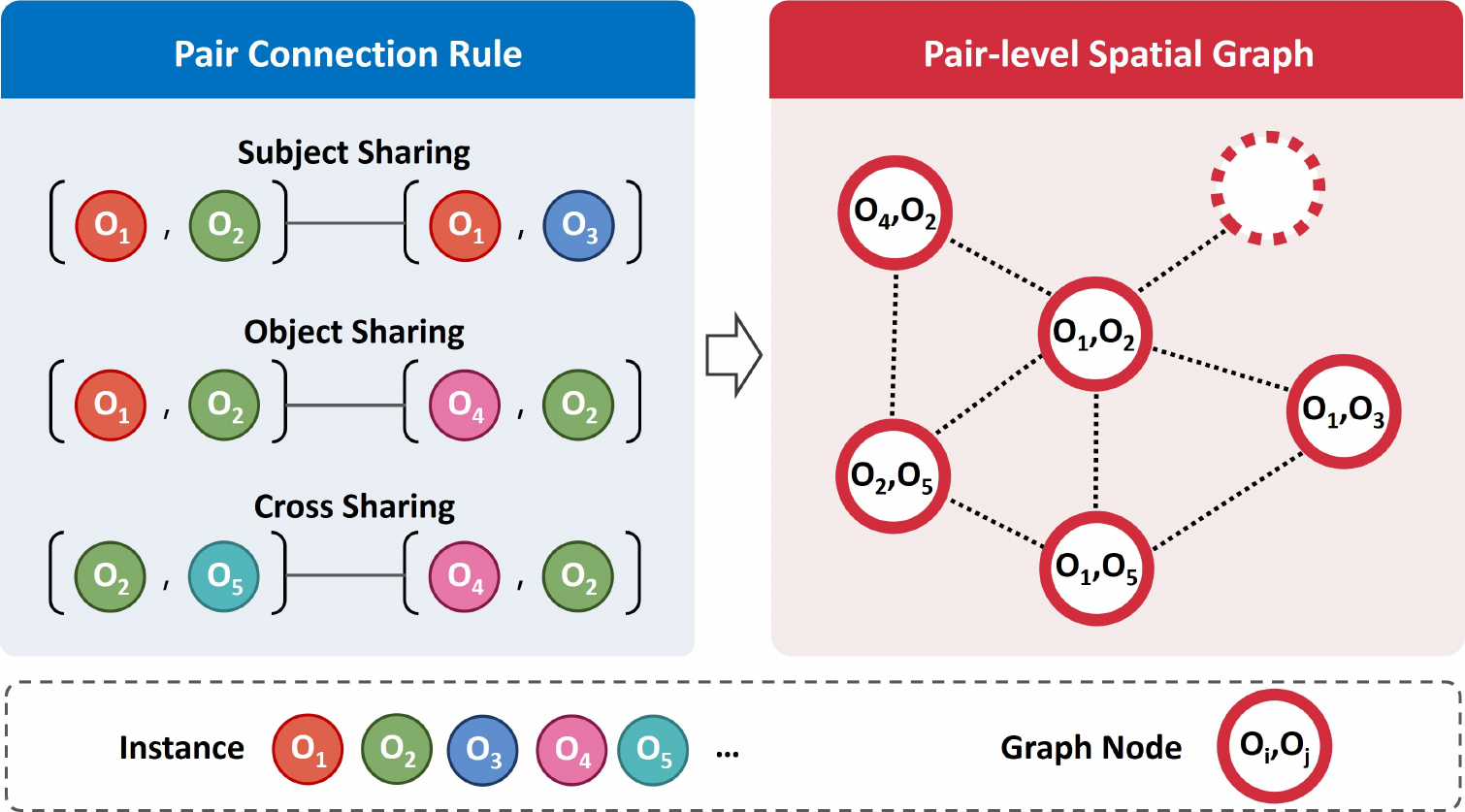}
	\end{center}
	\caption{Illustration of the spatial graph construction.}
	\label{fig:Pair-graph}
\end{figure}

\subsection{Stage I: Baseline Video Panoptic Parsing}

Since the focus of this work is not to redesign the parsing module, in Stage I, the baseline video panoptic parsing pipelines~\cite{yang2023panoptic}, including IPS+T and VPS, are retained to obtain panoptic instances for each frame. Specifically, each instance is represented by a segmentation mask, a category label, and a feature embedding. These parsing results are then used as input for the subsequent spatio-temporal relationship reasoning stage.

\subsection{Stage II: Spatio-Temporal Cooperative Learning for Relationship Prediction}


\subsubsection{Memory-Guided Matching}
To alleviate identity drift caused by small object size, weak texture, occlusion, and background clutter in satellite video, the Memory-Guided Matching (MGM) module is introduced to enhance the reliability and robustness of cross-frame instance associations. MGM does not replace the Stage-I parser; instead, it refines the temporal consistency of parsed instance associations, thereby providing more reliable trajectory features for subsequent relationship reasoning. During video processing, MGM maintains a global dynamic trajectory memory bank, where each trajectory's current state is represented as $(id, c, f_{\mathrm{last}}, m_{\mathrm{last}})$, with $id$ denoting the trajectory identifier, $c$ the object category, $f_{\mathrm{last}}$ the frame index of the most recent observation, and $m_{\mathrm{last}}$ the corresponding instance mask. This memory is continuously updated over time and serves as a global reference for cross-frame matching.

At frame $t$, current instances are matched to historical trajectories
under semantic and temporal consistency constraints: only trajectories
that have matching category labels and fall within a predefined temporal
window are considered. Candidate matches are ranked by mask-level overlap, and a greedy assignment ensures that each instance matches at most one trajectory while each trajectory is assigned at most once. After matching, successfully associated trajectories are updated in an overwrite manner, replacing the mask and timestamp with the current observation to reflect the latest instance state and prevent temporal accumulation of historical noise. Unmatched instances are appended to the memory as new trajectories, while unmatched historical trajectories remain in memory but are temporarily inactive. This design improves the temporal consistency of object tracking across frames, even under scale variations, shape changes, and local perturbations, while keeping computational complexity manageable.

\subsubsection{Spatial Context Enhancement}

In complex satellite video scenarios, instances and their interactions
form an interconnected and dynamically evolving system, where
dependencies exist not only between individual instances but also among
instance pairs. To capture such dependencies, high-value pairs are first
generated following the pair selection strategy~\cite{yang2023panoptic},
and a pair-level spatial graph is constructed according to the rules
illustrated in Fig.~\ref{fig:Pair-graph}. In this graph, each node
represents an instance pair, and edges are established between pair
nodes that share the same subject, share the same object, or exhibit a
cross-pair dependency, where the subject of one pair is identical to the
object of another pair, or vice versa. By explicitly modeling structured
dependencies among instance pairs, the spatial graph enables
relationship features to incorporate broader scene-level contextual
information, thereby enhancing the representation of complex object
interactions.

For each proposal pair $(s_q,o_q)$, the subject and object features at
frame $t$ are first encoded separately and concatenated to construct the
pair representation
$\mathbf{x}_q^t=
[\mathbf{f}_{s_q}^t;\mathbf{f}_{o_q}^t]$,
where $[\cdot;\cdot]$ denotes feature concatenation. The resulting pair
representation is used as the initial feature of the corresponding graph
node, i.e., $\mathbf{z}_q^{t,(0)}=\mathbf{x}_q^t$. Specifically, the subject and object encoders each produce a 256-dimensional embedding, resulting in a 512-dimensional pair representation after feature concatenation.

For each frame $t$, multi-layer message passing is performed independently
on the pair-level spatial graph. Let
$\mathbf{z}_q^{t,(\ell)}$ denote the representation of the $q$-th
node at layer $\ell$. The features of its neighboring nodes are
aggregated through mean pooling and transformed to obtain the neighborhood
representation $\mathbf{a}_q^{t,(\ell)}$. The neighborhood representation is then concatenated with the node's
own feature. Let
$\mathbf{m}_q^{t,(\ell)}
=
\Phi([\mathbf{z}_q^{t,(\ell)};
\mathbf{a}_q^{t,(\ell)}])$
denote the resulting update feature. The node representation is updated
through residual learning and layer normalization as
\begin{equation}
\mathbf{z}_q^{t,(\ell+1)}
=
\operatorname{ReLU}\!\left(
\operatorname{Norm}\!\left(
\mathbf{m}_q^{t,(\ell)}
+
\mathbf{z}_q^{t,(\ell)}
\right)\right),
\end{equation}
where $\operatorname{Norm}(\cdot)$ denotes layer normalization, and $\operatorname{ReLU}(\cdot)$ denotes the nonlinear activation function. Dropout is further applied to the updated node features during training. By iterating this process over multiple layers, spatial contextual information is propagated across the graph, enhancing semantic consistency among instance pairs and improving relationship representations in complex scenes.


\subsubsection{Multi-Scale Temporal Learning}
After capturing the spatial context of each instance pair via the SCE module, the corresponding relationship features are stacked along the temporal dimension to form a sequence
$\mathbf{Z}_q=\{\mathbf{z}_q^1,\mathbf{z}_q^2,\ldots,\mathbf{z}_q^T\}$.
Spatio-temporal relationships in satellite video often require sequential observations across multiple frames for reliable inference. To model such temporal dependencies, the MTL module employs an $L$-layer temporal convolutional network (TCN)~\cite{bai2018empirical} to encode the relationship feature sequence of each instance pair and capture temporal contextual information. Each TCN layer utilizes a one-dimensional dilated convolution, which progressively expands the temporal receptive field while preserving the temporal resolution. Specifically, the dilation rate of the $l$-th TCN layer is set to $d_l=2^{l-1}$ for $l=1,\ldots,L$. Let
$\widetilde{\mathbf{H}}_q^{(l)}
=\operatorname{Conv1D}_{d_l}(\mathbf{H}_q^{(l-1)})$
denote the intermediate output of the dilated convolution at the
$l$-th layer. The layer output is then computed as:
\begin{equation}
\mathbf{H}_q^{(l)}
=
\mathbf{H}_q^{(l-1)}
+
\operatorname{Drop}\!\left(
\sigma\!\left(
\operatorname{Norm}\!\left(\widetilde{\mathbf{H}}_q^{(l)}\right)
\right)\right),
\end{equation}
where $\mathbf{H}_q^{(0)}=\mathbf{Z}_q$,
$\operatorname{Conv1D}_{d_l}(\cdot)$ denotes a one-dimensional
convolution with dilation rate $d_l$,
$\operatorname{Norm}(\cdot)$ denotes a normalization layer,
$\sigma(\cdot)$ is the ReLU activation function, and
$\operatorname{Drop}(\cdot)$ denotes dropout. The residual connection
facilitates stable optimization and preserves the original relationship
information during multi-scale temporal modeling.

Finally, the temporally enhanced pair representations are fed into a shared prediction network with two complementary heads. The predicate head estimates video-level relationship scores by aggregating frame-wise responses over time, while the temporal span head predicts the frame-wise validity of each relationship. During inference, the relationship scores and frame-wise temporal predictions are jointly decoded into temporally grounded relationship triplets, which are subsequently ranked according to their confidence scores.

\subsubsection{Training Objective}
With the video parsing module fixed, the spatio-temporal relationship
prediction stage is optimized for proposal pair generation, predicate
classification, and temporal span prediction. The training objective is defined as:
\begin{equation}
\mathcal{L}
=
\lambda_{\mathrm{1}}\mathcal{L}_{\mathrm{pair}}
+
\lambda_{\mathrm{2}}\mathcal{L}_{\mathrm{pred}}
+
\lambda_{\mathrm{3}}\mathcal{L}_{\mathrm{span}},
\end{equation}
where $\mathcal{L}_{\mathrm{pair}}$, $\mathcal{L}_{\mathrm{pred}}$, and
$\mathcal{L}_{\mathrm{span}}$ denote the losses for proposal pair
generation, predicate classification, and frame-wise temporal span
prediction, respectively.

To alleviate the long-tailed distribution of predicate categories,
$\mathcal{L}_{\mathrm{pred}}$ is implemented using a class-reweighted
binary cross-entropy loss. A log-sum-exp-based multi-label ranking loss
is adopted for both $\mathcal{L}_{\mathrm{pair}}$ and
$\mathcal{L}_{\mathrm{span}}$, where the temporal span loss is computed only for a predicate of a proposal pair if that predicate is valid in at least one frame.

\begin{table*}[t]
\centering
\small
\caption{Comparison of TPSG baselines for the PredCls task (\%) on the T-STAR test set.}
\label{tab:predcls}
\renewcommand{\arraystretch}{1.68}
\resizebox{\textwidth}{!}{
\begin{tabular}{c|c c|c c c|c c c}
\hline
\multicolumn{3}{c|}{\textbf{Model}} &
\multicolumn{3}{c|}{\textbf{PredCls ($\theta_t=0.5$)}} &
\multicolumn{3}{c}{\textbf{PredCls ($\theta_t=0.1$)}} \\
\hline
\begin{tabular}[c]{@{}c@{}}Video\\Parsing\end{tabular} & \begin{tabular}[c]{@{}c@{}}Relationship\\Prediction\end{tabular} & \begin{tabular}[c]{@{}c@{}}Venue\end{tabular} &
\begin{tabular}[c]{@{}c@{}}\textbf{TR@}\\200/500/1000\end{tabular} &
\begin{tabular}[c]{@{}c@{}}\textbf{mTR@}\\200/500/1000\end{tabular} &
\begin{tabular}[c]{@{}c@{}}\textbf{FTR@}\\200/500/1000\end{tabular} &
\begin{tabular}[c]{@{}c@{}}\textbf{TR@}\\200/500/1000\end{tabular} &
\begin{tabular}[c]{@{}c@{}}\textbf{mTR@}\\200/500/1000\end{tabular} &
\begin{tabular}[c]{@{}c@{}}\textbf{FTR@}\\200/500/1000\end{tabular} \\
\hline
\multirow{7}{*}{VPS}
& Vanilla~\cite{yang2023panoptic} & CVPR'23 & 5.74/9.33/12.80 & 7.10/10.55/13.88 & 6.35/9.90/13.32 & 7.66/13.87/17.53 & 9.59/15.43/19.27 & 8.52/14.61/18.36 \\
& Filter~\cite{yang2023panoptic} & CVPR'23 & 4.27/6.99/8.34 & 5.02/7.13/7.62 & 4.61/7.06/7.96 & 9.49/16.37/20.08 & 11.22/16.72/18.24 & 10.28/16.54/19.12 \\
& Conv~\cite{yang2023panoptic} & CVPR'23 & 8.08/13.42/17.08 & 9.12/13.61/15.14 & 8.57/13.52/16.05 & 11.58/19.31/24.76 & 12.81/18.77/20.89 & 12.16/19.04/22.66 \\
& Transformer~\cite{yang2023panoptic} & CVPR'23 & 6.24/11.94/21.63 & 7.94/12.32/16.15 & 6.99/12.13/18.49 & 8.69/17.10/29.93 & 11.53/17.71/22.58 & 9.91/17.40/25.74 \\
& MCL~\cite{nguyen2025motion} & AAAI'25 & 4.31/7.00/11.09 & 5.25/7.60/9.93 & 4.73/7.29/10.48 & 6.34/10.78/16.39 & 7.92/12.15/15.71 & 7.04/11.42/16.04 \\
& IRG~\cite{li2025unbiased} & CVPR'25 & 8.79/17.00/23.72 & 9.26/15.54/18.06 & 9.02/16.24/20.51 & 10.60/21.46/33.12 & 11.61/19.50/23.83 & 11.09/20.44/27.72 \\
& STCL (Ours) & - & \textbf{16.71}/\textbf{29.38}/\textbf{41.34} & \textbf{17.76}/\textbf{23.64}/\textbf{26.95} & \textbf{17.22}/\textbf{26.20}/\textbf{32.63} & \textbf{18.12}/\textbf{31.33}/\textbf{43.60} & \textbf{20.88}/\textbf{26.90}/\textbf{30.27} & \textbf{19.40}/\textbf{28.95}/\textbf{35.73} \\
\hline
\multirow{7}{*}{IPS+T}
& Vanilla~\cite{yang2023panoptic} & CVPR'23 & 24.99/37.19/47.83 & 23.57/30.13/33.87 & 24.26/33.29/39.66 & 25.44/37.90/48.74 & 24.55/31.55/35.67 & 24.99/34.43/41.19 \\
& Filter~\cite{yang2023panoptic} & CVPR'23 & 30.99/42.12/53.33 & 27.20/33.62/37.26 & 28.97/37.39/43.87 & 31.53/42.82/54.16 & 29.23/36.09/40.09 & 30.34/39.17/46.07 \\
& Conv~\cite{yang2023panoptic} & CVPR'23 & 32.15/44.02/55.65 & 29.37/35.34/38.14 & 30.69/39.21/45.26 & 32.52/44.56/56.23 & 29.78/35.92/38.79 & 31.09/39.78/45.91 \\
& Transformer~\cite{yang2023panoptic} & CVPR'23 & 30.33/43.32/54.32 & 29.79/35.80/39.00 & 30.05/39.20/45.40 & 30.78/43.98/55.07 & 31.30/37.66/41.07 & 31.04/40.58/47.05 \\
& MCL~\cite{nguyen2025motion} & AAAI'25 & 30.99/44.23/53.37 & 28.17/34.27/37.32 & 29.51/38.62/43.93 & 31.49/44.85/53.99 & 30.51/37.00/40.05 & 30.99/40.55/45.99 \\
& IRG~\cite{li2025unbiased} & CVPR'25 & 30.45/47.25/60.07 & 21.65/28.67/33.02 & 25.30/35.68/42.61 & 31.07/48.03/\textbf{60.98} & 23.18/30.51/35.51 & 26.55/37.32/44.88 \\
& STCL (Ours) & - & \textbf{38.11}/\textbf{51.72}/\textbf{60.24} & \textbf{32.03}/\textbf{37.36}/\textbf{39.59} & \textbf{34.81}/\textbf{43.38}/\textbf{47.78} & \textbf{38.56}/\textbf{52.34}/60.94 & \textbf{34.32}/\textbf{39.79}/\textbf{42.19} & \textbf{36.32}/\textbf{45.21}/\textbf{49.86} \\
\hline
\end{tabular}}
\end{table*}

\section{Experiments}

\subsection{Experimental Setup}

\textbf{Evaluation Tasks and Metrics.} We adopt the standard evaluation protocol for spatio-temporal scene graph generation and report Temporal Recall@K (TR@K), mean Temporal Recall@K (mTR@K), and their harmonic mean FTR@K. TR@K measures the overall recall of ground-truth spatio-temporal triplets within the top-$K$ predictions. mTR@K averages recall across relationship categories to alleviate the influence of class imbalance. FTR@K provides a balanced summary of overall recall and category-level fairness. Since the number of potential relationships in satellite video is substantially larger than that in natural videos, we set $K$ to 200, 500, and 1000 throughout the experiments.

In accordance with the conventional spatio-temporal SGG task, the TPSG task is divided into two sub-tasks. Predicate Classification (PredCls): ground-truth instance trajectories and categories are provided, while visual trajectory features are extracted using the corresponding video parsing pipeline to ensure a fair comparison. Scene Graph Detection (SGDet): object trajectories, categories, and trajectory features are inferred from the input video sequence, while the downstream spatio-temporal relationship prediction pipeline remains unchanged. Following the strict evaluation protocol~\cite{yang2023panoptic}, a predicted triplet is considered correct only if all of the following criteria are simultaneously satisfied: 1) the subject, object, and relationship categories are accurately predicted; 2) the temporal overlap between the predicted and ground-truth intervals meets or exceeds a predefined threshold; and 3) within the temporal intersection, the per-frame mask IoU for both subject and object exceeds the spatial threshold. Unless otherwise specified, the spatial threshold is fixed at 0.5, and results are reported under two temporal thresholds, i.e., $\theta_t=0.5$ and $\theta_t=0.1$, to evaluate both strict and relaxed temporal localization performance.

\textbf{Baseline Models.} For Stage I video panoptic parsing, we adopt two baseline methods~\cite{yang2023panoptic}: \textit{IPS+T} and \textit{VPS}. For Stage II spatio-temporal relationship prediction, the proposed STCL is compared with six representative methods: \textit{1) Fully Connected Layer (Vanilla)}~\cite{yang2023panoptic}: The fused pair-level features are processed through a standard fully connected layer. \textit{2) Handcrafted Temporal Filter (Filter)}~\cite{yang2023panoptic}: Local temporal dependencies are captured via a fixed kernel aggregating features from neighboring frames. \textit{3) 1D Convolutional Layer (Conv)}~\cite{yang2023panoptic}: A learnable 1D convolution replaces the handcrafted filter to adaptively model temporal dependencies. \textit{4) Transformer Encoder (Transformer)}~\cite{yang2023panoptic}: A 3-layer transformer block with positional embeddings is applied to the fused pair features. \textit{5) Motion-aware Contrastive Learning (MCL)}~\cite{nguyen2025motion}: Motion-sensitive relationship representations are learned by constructing positive and negative trajectory pair samples. \textit{6) Iterative Relationship Generator (IRG)}~\cite{li2025unbiased}: Relationship features of object pairs are iteratively refined through contextual propagation. To ensure a fair comparison, all methods share the same Stage I parsing setup and pair selection strategy, differing only in the Stage II spatio-temporal relationship prediction modules.

\textbf{Implementation Details.} All experiments are conducted on the T-STAR dataset, which is split at the video level into training, validation, and test sets in a 6:2:2 ratio. To ensure a fair comparison, the spatio-temporal relationship prediction models are trained on top of a frozen video panoptic parsing network using the training set and evaluated on the test set. Model optimization is performed using the Adam~\cite{kingma2014adam} optimizer with an initial learning rate of $1\times10^{-4}$. The TCN comprises four layers with a kernel size of 3 and dilation rates of 1, 2, 4, and 8. Group normalization and dropout with a rate of 0.1 are applied in each temporal convolutional block. The loss weights are set to $\lambda_{1}=1$,
$\lambda_{2}=5$, and $\lambda_{3}=1$. All experiments are carried out on a system running Ubuntu 18.04.6 LTS with a single NVIDIA TITAN RTX GPU. For reproducibility, the random seed is fixed at 42.

\subsection{Comparison with Representative Baseline Methods}
To comprehensively evaluate the performance of the proposed method for the TPSG task in satellite video, we conduct experiments with two video panoptic parsing pipelines, i.e., IPS+T and VPS, and compare STCL with six representative spatio-temporal relationship prediction methods, including Vanilla, Filter, Conv, Transformer, MCL, and IRG. Given that temporal interval prediction can significantly affect evaluation outcomes, we report results under two temporal thresholds for both PredCls and SGDet tasks in Tables~\ref{tab:predcls} and~\ref{tab:sgdet}, respectively. Additionally, Table~\ref{tab:time} presents the performance and inference efficiency of different methods on SGDet.

As shown in Table~\ref{tab:predcls}, under both temporal thresholds, STCL achieves the strongest overall performance across both IPS+T and VPS settings, obtaining the best results on most metrics. This indicates that STCL effectively leverages trajectory-level spatial and temporal context to robustly model spatio-temporal relationship semantics. Among the baselines, Vanilla and Filter generally show weaker performance, suggesting that simple feature mapping or fixed temporal filtering is insufficient to capture complex relationship evolution. Conv and Transformer improve performance in several settings by introducing learnable temporal modeling or global contextual aggregation, while their gains vary across parsing pathways and evaluation metrics. MCL achieves competitive results on certain metrics but shows noticeable performance variations between IPS+T and VPS. IRG further enhances relationship representation through contextual propagation, achieving particularly competitive TR@1000 results under the IPS+T pathway. Nevertheless, these methods mainly focus on local temporal modeling or partial relationship refinement, and are less effective in jointly handling identity consistency, spatial structure, and temporal evolution in satellite video. Notably, STCL with IPS+T substantially outperforms that with VPS; for instance, under the temporal threshold of 0.5, FTR@200/500/1000 reaches 34.81/43.38/47.78 on IPS+T, compared with 17.22/26.20/32.63 on VPS. This underscores the importance of stable trajectory representations for accurate spatio-temporal relationship prediction.

\begin{table*}[t]
\centering
\small
\caption{Comparison of TPSG baselines for the SGDet task (\%) on the T-STAR test set.}
\label{tab:sgdet}
\renewcommand{\arraystretch}{1.68}
\resizebox{\textwidth}{!}{
\begin{tabular}{c|c c|c c c|c c c}
\hline
\multicolumn{3}{c|}{\textbf{Model}} &
\multicolumn{3}{c|}{\textbf{SGDet  ($\theta_t=0.5$)}} &
\multicolumn{3}{c}{\textbf{SGDet ($\theta_t=0.1$)}} \\
\hline
\begin{tabular}[c]{@{}c@{}}Video\\Parsing\end{tabular} & \begin{tabular}[c]{@{}c@{}}Relationship\\Prediction\end{tabular} & \begin{tabular}[c]{@{}c@{}}Venue\end{tabular} &
\begin{tabular}[c]{@{}c@{}}\textbf{TR@}\\200/500/1000\end{tabular} &
\begin{tabular}[c]{@{}c@{}}\textbf{mTR@}\\200/500/1000\end{tabular} &
\begin{tabular}[c]{@{}c@{}}\textbf{FTR@}\\200/500/1000\end{tabular} &
\begin{tabular}[c]{@{}c@{}}\textbf{TR@}\\200/500/1000\end{tabular} &
\begin{tabular}[c]{@{}c@{}}\textbf{mTR@}\\200/500/1000\end{tabular} &
\begin{tabular}[c]{@{}c@{}}\textbf{FTR@}\\200/500/1000\end{tabular} \\
\hline
\multirow{7}{*}{VPS}
& Vanilla~\cite{yang2023panoptic} & CVPR'23
& 9.32/14.00/17.60 & 10.54/13.34/15.15 & 9.90/13.66/16.29
& 12.60/19.93/25.56 & 14.04/18.15/21.19 & 13.28/19.00/23.17 \\
& Filter~\cite{yang2023panoptic} & CVPR'23
& 5.93/9.42/10.83 & 5.88/7.91/8.45 & 5.90/8.60/9.49
& 13.32/22.39/25.97 & 13.64/18.79/20.54 & 13.48/20.43/22.93 \\
& Conv~\cite{yang2023panoptic} & CVPR'23
& 10.56/16.85/20.48 & 9.76/13.62/15.12 & 10.14/15.06/17.40
& 14.82/24.72/30.16 & 14.01/19.81/22.16 & 14.41/22.00/25.55 \\
& Transformer~\cite{yang2023panoptic} & CVPR'23
& 8.43/15.10/21.82 & 8.48/13.45/16.48 & 8.45/14.23/18.78
& 11.61/22.15/32.85 & 11.44/18.49/23.52 & 11.53/20.15/27.41 \\
& MCL~\cite{nguyen2025motion} & AAAI'25
& 6.81/10.29/12.49 & 7.02/9.43/10.51 & 6.92/9.84/11.42
& 9.39/14.57/18.08 & 9.36/12.61/14.46 & 9.37/13.52/16.07 \\
& IRG~\cite{li2025unbiased} & CVPR'25
& 11.17/18.80/24.39 & 10.92/15.49/18.22 & 11.05/16.99/20.86
& 14.42/25.36/36.04 & 14.99/21.72/27.55 & 14.70/23.40/31.23 \\
& STCL (Ours) & -
& \textbf{22.34}/\textbf{32.09}/\textbf{40.08} & \textbf{20.31}/\textbf{25.48}/\textbf{28.21} & \textbf{21.28}/\textbf{28.41}/\textbf{33.12} & \textbf{24.55}/\textbf{35.33}/\textbf{44.26} & \textbf{22.35}/\textbf{27.76}/\textbf{30.71} & \textbf{23.40}/\textbf{31.09}/\textbf{36.26} \\
\hline
\multirow{7}{*}{IPS+T}
& Vanilla~\cite{yang2023panoptic} & CVPR'23
& 10.53/13.19/15.80 & 11.33/12.38/13.11 & 10.92/12.77/14.33
& 13.75/16.91/19.96 & 16.27/17.64/20.09 & 14.90/17.27/20.02 \\
& Filter~\cite{yang2023panoptic} & CVPR'23
& 18.63/23.17/23.67 & 16.07/17.18/17.32 & 17.26/19.73/20.00
& 22.73/28.55/29.32 & 21.44/22.99/23.21 & 22.06/25.47/25.91 \\
& Conv~\cite{yang2023panoptic} & CVPR'23
& 21.45/26.33/27.49 & 17.76/19.72/19.86 & 19.43/22.55/23.06
& 24.00/29.88/31.10 & 21.42/24.94/25.09 & 22.64/27.19/27.77 \\
& Transformer~\cite{yang2023panoptic} & CVPR'23
& 21.62/26.44/27.49 & 18.85/20.10/20.51 & 20.14/22.84/23.49
& 24.33/29.77/30.99 & 22.80/24.36/24.79 & 23.54/26.79/27.54 \\
& MCL~\cite{nguyen2025motion} & AAAI'25
& 15.96/20.07/25.78 & 13.85/17.01/19.30 & 14.83/18.41/22.07
& 17.85/22.51/28.66 & 17.42/20.98/23.36 & 17.63/21.72/25.74 \\
& IRG~\cite{li2025unbiased} & CVPR'25
& 23.89/30.60/31.54 & 16.03/17.98/18.16 & 19.18/22.65/23.05
& 27.05/34.87/36.36 & 21.45/24.20/24.50 & 23.93/28.57/29.27 \\
& STCL (Ours) & -
& \textbf{28.21}/\textbf{32.37}/\textbf{34.37} & \textbf{22.96}/\textbf{24.61}/\textbf{25.27} & \textbf{25.32}/\textbf{27.96}/\textbf{29.13}
& \textbf{31.54}/\textbf{36.04}/\textbf{38.59} & \textbf{28.38}/\textbf{30.06}/\textbf{31.16} & \textbf{29.88}/\textbf{32.78}/\textbf{34.48} \\
\hline
\end{tabular}}
\end{table*}

Compared with PredCls, SGDet requires models to generate instance masks, categories, and trajectory features directly from raw satellite video inputs, thereby introducing additional parsing noise and error propagation. Consequently, SGDet evaluates not only the effectiveness of spatio-temporal relationship modeling, but also the quality of instance segmentation and cross-frame association. As shown in Table~\ref{tab:sgdet}, STCL consistently achieves the best results across all configurations, demonstrating its robustness under model-generated trajectories and noisy relationship inputs. The relative rankings of the baseline methods vary across parsing pathways, temporal thresholds, and top-K settings. In general, Transformer, Conv, and IRG achieve competitive results among the baselines, suggesting that both temporal dependency modeling and contextual relationship refinement are beneficial for SGDet. However, their performance remains limited when dealing with trajectory noise and relationship jitter. By jointly enhancing memory-guided instance consistency, pair-level spatial context, and multi-scale temporal modeling, STCL provides more robust spatio-temporal relationship reasoning under the challenging SGDet setting.

Table~\ref{tab:time} compares the predictive performance and inference
efficiency of different relationship prediction methods under the SGDet
setting using a single NVIDIA TITAN RTX GPU. FPS denotes the
single-device inference throughput in frames per second, while
Time/Video denotes the average single-device inference time per video
in seconds. The reported inference time covers only Stage II
relationship prediction. STCL achieves the best FTR@200/500/1000,
substantially outperforming Vanilla, Filter, Conv, Transformer, MCL,
and IRG. This improvement demonstrates the effectiveness of integrating
memory-guided matching with spatial and temporal context modeling for
complex relationship reasoning in satellite video. Although STCL has a
slightly lower frame rate and a longer per-video inference time than
some lightweight baselines, it still achieves a throughput of
114.09 FPS and an average inference time of 1.96 s per video. Overall,
STCL provides substantial accuracy improvements while maintaining
practical inference efficiency.

\begin{figure*}[!tb]
	\begin{center}
		\includegraphics[width=0.98\linewidth]{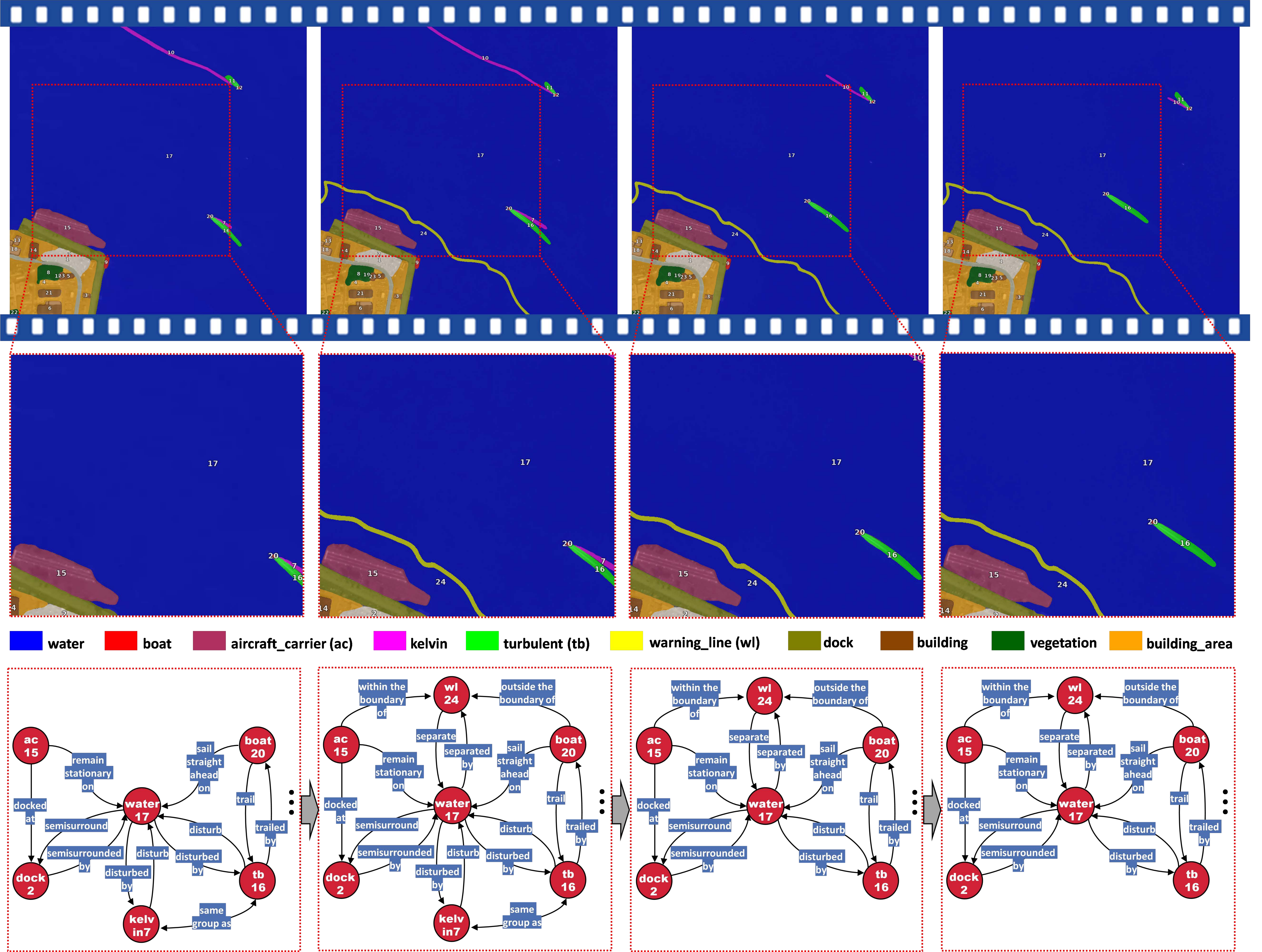}
	\end{center}
	\caption{Qualitative visualization of STCL in a port scene.}
	\label{fig:port}
\end{figure*}

\begin{figure*}[htbp]
	\begin{center}
		\includegraphics[width=0.98\linewidth]{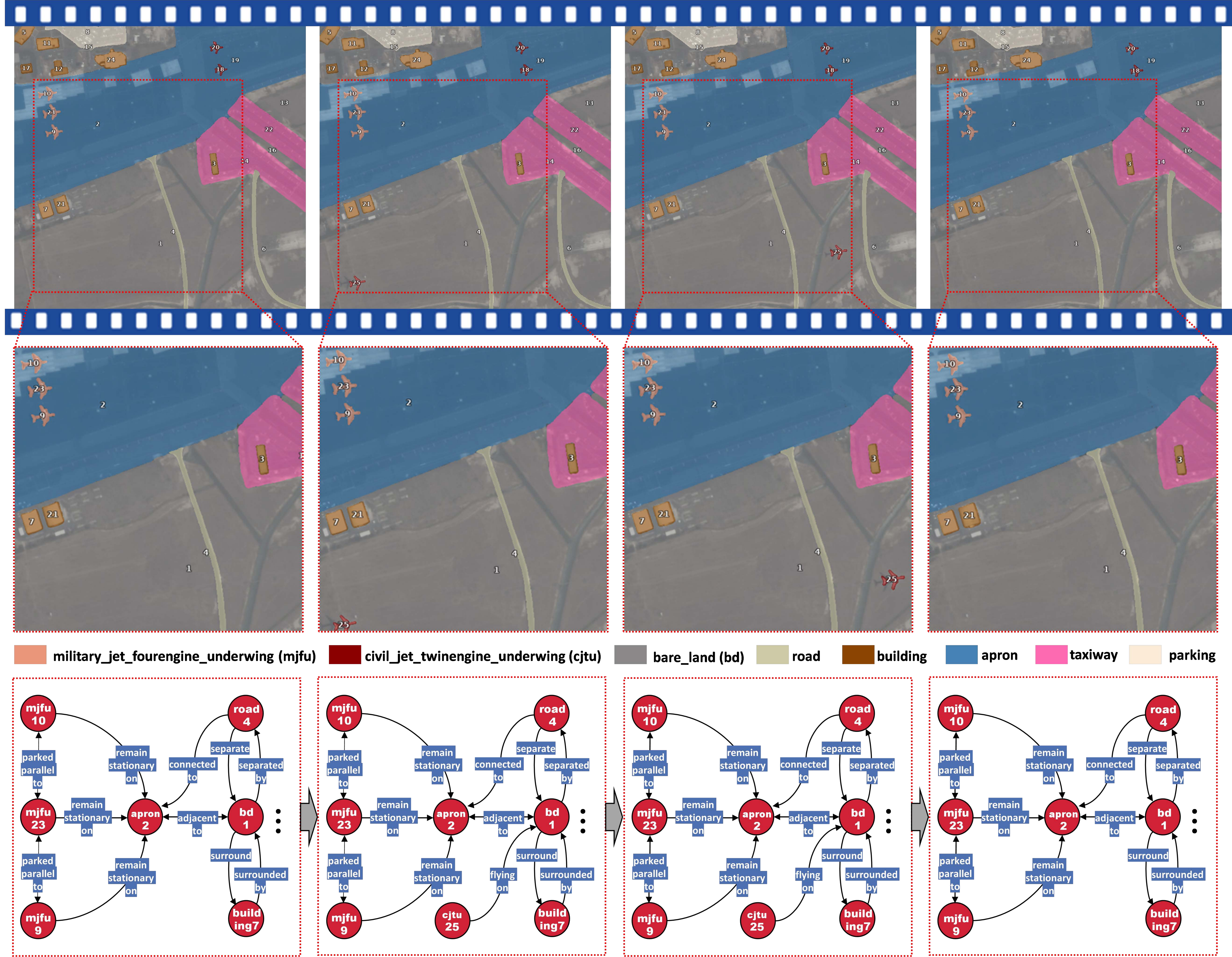}
	\end{center}
	\caption{Qualitative visualization of STCL in an airport scene.}
	\label{fig:airport}
\end{figure*}

\subsection{Qualitative Results and Visualization}

To qualitatively assess the spatio-temporal relationship modeling capabilities of the proposed STCL method in typical satellite video scenarios, we visualize the generated spatio-temporal scene graphs. Fig.~\ref{fig:port} and Fig.~\ref{fig:airport} present representative visualizations in port and airport scenes, respectively. These results provide an intuitive demonstration of the model’s ability to capture spatial structure, multi-object interactions, and temporal evolution of relationships.

As illustrated in Fig.~\ref{fig:port}, in the port scenario, STCL preserves instance identity across frames while accurately capturing the temporal evolution of multi-object interactions. For example, the model correctly identifies spatio-temporal triplets such as \textit{$<$aircraft\_carrier15, docked at, dock2$>$} and \textit{$<$aircraft\_carrier15, remain stationary on, water17$>$}, reflecting the persistence of relationships over time. Moreover, the spatial context modeling enables precise recognition of triplets such as \textit{$<$aircraft\_carrier15, within the boundary of, warning\_line24$>$} and \textit{$<$boat20, outside the boundary of, warning\_line24$>$}, demonstrating high-fidelity understanding of spatial interactions. In addition, leveraging cross-object interactions, STCL successfully infers complex triplets like \textit{$<$turbulent16, same group as, kelvin7$>$}, highlighting its sensitivity and reasoning capability in multi-object dynamic scenarios.

\begin{table}[t]
\centering
\small
\caption{Performance and inference efficiency comparison on the IPS+T-based SGDet task using a single NVIDIA TITAN RTX GPU.}
\label{tab:time}
\renewcommand{\arraystretch}{1.6}
\resizebox{0.92\columnwidth}{!}{
\begin{tabular}{c|ccc}
\hline
\multirow{2}{*}{\raisebox{-2.9ex}{\textbf{Model}}}& \multicolumn{3}{c}{\textbf{SGDet ($\theta_t=0.5$)}} \\
\cline{2-4}
& \begin{tabular}[c]{@{}c@{}}\textbf{FTR@}\\200/500/1000\end{tabular} & \textbf{FPS} & \begin{tabular}[c]{@{}c@{}}\textbf{Time/Video} (s)\end{tabular} \\
\hline
Vanilla~\cite{yang2023panoptic} & 10.92/12.77/14.33 & 140.83 & 1.59 \\
Filter~\cite{yang2023panoptic} & 17.26/19.73/20.00 & 142.76 & 1.57 \\
Conv~\cite{yang2023panoptic} & 19.43/22.55/23.06 & 139.92 & 1.60 \\
Transformer~\cite{yang2023panoptic} & 20.14/22.84/23.49 & 136.27 & 1.64 \\
MCL~\cite{nguyen2025motion} & 14.83/18.41/22.07 & \textbf{147.86} & \textbf{1.51} \\
IRG~\cite{li2025unbiased} & 19.18/22.65/23.05 & 132.15 & 1.69 \\
\textbf{STCL (Ours)} & \textbf{25.32}/\textbf{27.96}/\textbf{29.13} & 114.09 & 1.96 \\
\hline
\end{tabular}}
\end{table}

Fig.~\ref{fig:airport} presents qualitative results of STCL in an airport scene. STCL accurately captures spatial relationships such as 
\textit{$<$bare\_land1, surround, building7$>$}, 
reflecting its ability to model the structural organization of complex scenes. It further recognizes the \textit{remain stationary on} relationship
between airplanes and runways, as well as the \textit{parked parallel to}
relationship between different airplanes. These results demonstrate that STCL effectively integrates identity-consistent instance trajectories, pair-level spatial context, and temporal dependencies to reason about evolving inter-object relationships beyond single-frame spatial observations.

\begin{table}[t]
\centering
\small
\caption{Ablation study on key components of STCL.}
\label{tab:component_ablation}
\renewcommand{\arraystretch}{2.0}
\setlength{\tabcolsep}{3pt}
\resizebox{0.98\columnwidth}{!}{
\begin{tabular}{ccc|ccc}
\hline
\multirow{2}{*}[-1.8ex]{\textbf{MGM}} & 
\multirow{2}{*}[-1.8ex]{\textbf{SCE}} & 
\multirow{2}{*}[-1.8ex]{\textbf{MTL}} &
\multicolumn{3}{c}{\textbf{SGDet} ($\theta_t=0.5$)} \\
\cline{4-6}
 &  &  & 
\begin{tabular}[c]{@{}c@{}}\textbf{TR@}\\200/500/1000\end{tabular} & 
\begin{tabular}[c]{@{}c@{}}\textbf{mTR@}\\200/500/1000\end{tabular} & 
\begin{tabular}[c]{@{}c@{}}\textbf{FTR@}\\200/500/1000\end{tabular} \\
\hline
 & $\checkmark$ & $\checkmark$ & 27.88/31.43/33.54 & 22.01/23.45/24.15 & 24.60/26.86/28.08 \\
$\checkmark$ & $\checkmark$ &  & 11.83/15.16/17.43 & 11.75/13.69/14.25 & 11.79/14.39/15.68 \\
$\checkmark$ &  & $\checkmark$ & 26.10/31.32/33.09 & 20.40/23.38/24.03 & 22.90/26.77/27.84 \\
$\checkmark$ & $\checkmark$ & $\checkmark$ & 
\textbf{28.21}/\textbf{32.37}/\textbf{34.37} & 
\textbf{22.96}/\textbf{24.61}/\textbf{25.27} & 
\textbf{25.32}/\textbf{27.96}/\textbf{29.13} \\
\hline
\end{tabular}}
\end{table}

\subsection{Ablation Studies}

To further understand the effectiveness of the proposed STCL framework, we conduct ablation studies from two perspectives. First, we evaluate the contribution of each key component, including memory-guided matching (MGM), spatial context enhancement (SCE), and multi-scale temporal learning (MTL). Since MTL is observed to play the most critical role in spatio-temporal relationship prediction, we further analyze the influence of its temporal modeling depth. All ablation experiments are conducted under the IPS+T-based SGDet setting and evaluated using TR@K, mTR@K, and FTR@K.

\textbf{Effect of STCL Components.}
To assess the contribution of each key component in STCL, we conduct
ablation experiments by removing MGM, SCE, or MTL individually from the
complete model. As shown in Table~\ref{tab:component_ablation}, all three components contribute positively to spatio-temporal relationship prediction, while their effects differ in magnitude. Among them, MTL plays the most critical role. Removing MTL causes a substantial drop in TR@200/500/1000, mTR@200/500/1000, and FTR@200/500/1000, demonstrating that dynamic temporal modeling is essential for recognizing spatio-temporal relationships. This is because relationships in instance-level satellite video are highly dependent on object motion patterns, relationship duration, and temporal state evolution. By modeling relationship features over multiple temporal scales, MTL effectively captures both short-term variations and long-range dependencies. Removing SCE or MGM also degrades performance, but the decrease is relatively smaller than that caused by removing MTL. This indicates that SCE improves the discriminability of relationship representations by modeling structured dependencies among different instance pairs within the same frame, whereas MGM enhances the stability of cross-frame instance association and provides more reliable trajectory foundations for subsequent spatio-temporal relationship modeling. When all three components are jointly used, STCL achieves the best performance. These results verify the complementarity of the proposed modules: MGM provides temporally consistent instance trajectories, SCE enhances relationship features through intra-frame structural context, and MTL captures the dynamic evolution of relationships over time. Their collaboration enables STCL to model complex spatio-temporal relationships in satellite video more accurately and robustly.

\begin{table}[t]
\centering
\small
\caption{Effect of temporal modeling depth $L$ in the MTL module.}
\label{tab:depth_ablation}
\renewcommand{\arraystretch}{1.6}
\setlength{\tabcolsep}{8pt}
\resizebox{0.98\columnwidth}{!}{
\begin{tabular}{c|ccc}
\hline
\multirow{2}{*}{\raisebox{-2.2ex}{\textbf{$L$}}} & \multicolumn{3}{c}{\textbf{SGDet ($\theta_t=0.5$)}} \\
\cline{2-4}
& \begin{tabular}[c]{@{}c@{}}\textbf{TR@}\\200/500/1000\end{tabular}
& \begin{tabular}[c]{@{}c@{}}\textbf{mTR@}\\200/500/1000\end{tabular}
& \begin{tabular}[c]{@{}c@{}}\textbf{FTR@}\\200/500/1000\end{tabular} \\
\hline
1 & 26.26/31.87/33.20 & 20.04/21.65/22.07 & 22.73/25.79/26.52 \\
2 & 25.21/30.48/33.37 & 21.49/23.06/24.23 & 23.20/26.26/28.08 \\
3 & 25.76/30.98/33.09 & 20.41/22.50/23.21 & 22.78/26.07/27.29 \\
4 & \textbf{28.21}/\textbf{32.37}/\textbf{34.37} & \textbf{22.96}/\textbf{24.61}/\textbf{25.27} & \textbf{25.32}/\textbf{27.96}/\textbf{29.13} \\
5 & 25.60/30.54/32.59 & 18.91/21.57/22.33 & 21.75/25.28/26.50 \\
6 & 26.93/31.37/33.04 & 22.60/23.92/24.52 & 24.57/27.14/28.15 \\
7 & 27.60/31.59/32.82 & 21.43/23.51/24.03 & 24.12/26.96/27.74 \\
\hline
\end{tabular}}
\end{table}

\textbf{Effect of Temporal Modeling Depth.}
Given the dominant contribution of MTL, we further investigate the influence of temporal modeling depth on spatio-temporal relationship prediction by varying the number of TCN layers while keeping the spatial modeling configuration fixed. Specifically, the temporal modeling depth is set to 1, 2, 3, 4, 5, 6, and 7 layers. As shown in Table~\ref{tab:depth_ablation}, the temporal modeling depth has a clear influence on overall performance, with the best results achieved when $L$ is set to 4. This indicates that a properly enlarged temporal receptive field helps capture more reliable temporal dependencies from consecutive frames and thus benefits complex spatio-temporal relationship prediction. This trend is consistent with the nature of satellite video relationships, where relationship semantics often depend on object motion trends, relative geometric evolution, and temporal state transitions rather than isolated single-frame observations. For example, relationships such as \textit{accelerate along}, \textit{take off from}, \textit{approach}, and \textit{away from} generally require observations across multiple consecutive frames to be reliably inferred. However, when the temporal depth exceeds four layers, the performance generally remains below that of the four-layer setting, despite minor fluctuations, indicating that increasing the temporal modeling depth does not necessarily lead to further performance gains. One possible explanation is that excessive temporal depth may introduce over-smoothing along the temporal dimension, weakening local but discriminative motion cues; it may also increase model complexity and optimization difficulty, leading to redundant temporal modeling or overfitting under limited training data. In our experiments, the 4-layer TCN achieves the best trade-off between temporal context modeling capability and model complexity, and is therefore adopted as the default setting.


\section{Conclusion}
This paper presents T-STAR, the first large-scale benchmark dedicated to the TPSG task in satellite video. T-STAR contains satellite video frames with image sizes ranging from $400\times400$ to $1{,}500\times1{,}500$ pixels and provides more than 1.1 million instance masks and more than 3.8 million spatio-temporal
triplets. With fine-grained panoptic mask annotations, temporally consistent instance identities, and rich spatio-temporal relationships, T-STAR provides an essential data foundation for advancing TPSG in satellite video as a challenging and meaningful task. Furthermore, considering the unique characteristics of satellite video, we design an STCL model within a unified framework for the TPSG task to enhance cross-frame instance consistency and spatio-temporal relationship prediction. Future work will focus on extending T-STAR to more diverse satellite video scenarios and further exploring open-world TPSG, thereby advancing satellite video interpretation toward more general and scalable structured geospatial cognition.

\section*{Acknowledgments}
This work was partly supported by the Fundamental and Interdisciplinary Disciplines Breakthrough Plan of the Ministry of Education of China under Grant JYB2025XDXM111, the National Natural Science Foundation of China under Grant 42371321, and the Key Research and Development Program of Hubei Province under Grant 2025BAB024.

\ifCLASSOPTIONcaptionsoff
  \newpage
\fi

\bibliographystyle{IEEEtran}
\bibliography{IEEEegbib}
\vspace{-6mm}

\end{document}